\documentclass{article}

\usepackage{caption}
\usepackage{tabularx, colortbl}
\usepackage{float}
\usepackage{calc}

\newcolumntype{L}{>{\raggedright\arraybackslash}X}
\newcolumntype{R}{>{\raggedleft\arraybackslash}X}
\newcolumntype{C}{>{\centering\arraybackslash}X}

\usepackage{arxiv}

\usepackage[utf8]{inputenc} 
\usepackage[T1]{fontenc}    
\usepackage[colorlinks=true,linkcolor=blue,citecolor=blue]{hyperref}       
\usepackage{url}            
\usepackage{booktabs}       
\usepackage{amsfonts}       
\usepackage{nicefrac}       
\usepackage{microtype}      
\usepackage{lipsum}		
\usepackage{graphicx}
\usepackage[numbers]{natbib}
\usepackage{doi}

\renewcommand{\headeright}{}
\renewcommand{\undertitle}{}
\renewcommand{\shorttitle}{\emph{ChainPoll}: A High Efficacy Method for LLM Hallucination Detection}

\title{\emph{ChainPoll}: A High Efficacy Method for LLM Hallucination Detection}

\author{ Robert Friel\\
	Galileo Technologies Inc.\\
	\And
    Atindriyo Sanyal \\
    Galileo Technologies Inc.\\
}

\usepackage{setspace} 

\usepackage{etoolbox}
\AtBeginEnvironment{quote}{\par\singlespacing\small}

\usepackage[raggedrightboxes]{ragged2e}

\newcolumntype{b}{X}
\newcolumntype{s}{>{\hsize=.4\hsize}X}

\usepackage{pgfmath}

\newcommand\mybar[5]{
#1 \ {\color[RGB]{#3, #4, #5}\rule{#1\linewidth}{8pt}}}
  
\newcommand\mybarbold[5]{
\textbf{#1} \ {\color[RGB]{#3, #4, #5}\rule{#1\linewidth}{8pt}}}

\begin{document}

\maketitle








\begin{abstract}
Large language models (LLMs) have witnessed significant advancements in generating coherent, intelligent, and contextually relevant responses. However, the presence of hallucinations -- inaccurate or unmotivated claims -- remains a persistent challenge, motivating the development of automated metrics for the detection of hallucinations in LLM outputs.

We make two contributions: \emph{ChainPoll}, a novel hallucination detection methodology that substantially outperforms existing alternatives, and \emph{RealHall}, a carefully curated suite of benchmark datasets for evaluating hallucination detection metrics proposed in recent literature.

To construct \emph{RealHall}, we critically review tasks and datasets used in prior work on hallucination detection, finding that many of them have very limited relevance to the powerful LLMs used in practice today. To get rid of this limitation, we select four datasets that are truly challenging for state-of-the-art (modern era) LLMs and relevant to real world applications.

We use \emph{RealHall} to perform a head-to-head and non-biased comparison between \textit{ChainPoll} and a wide range of hallucination metrics proposed in recent literature and showcase that \textit{ChainPoll} achieves superior performance across all four of the benchmarks in \emph{RealHall}, with an aggregate AUROC of 0.781, beating the next best theoretical algorithm by 11\%, and beating industry standards for LLMs by over 23\%, while simultaneously being cheaper to compute and significantly more explainable than alternative metrics.

We propose 2 new metrics to quantify LLM hallucinations - \textit{Adherence} and \textit{Correctness}. The former pertinent to Retrieval Augmented Generation (RAG) workflows measuring an LLM's reasoning abilities within the provided documents and context, while the latter focused capturing general logical and reasoning based mistakes.

\end{abstract}


\vspace{10pt}

\tableofcontents

\section{Introduction}\label{Sec:Introduction}
\subsection{Summary of contributions}

Large language models (LLMs) have witnessed significant advancements in generating coherent, intelligent, and contextually relevant responses. However, the presence of hallucinations -- inaccurate or unmotivated claims -- remains a persistent challenge, motivating the development of automated metrics for the detection of hallucinations in LLM outputs.

This paper presents the research behind the Galileo platform's \textbf{state-of-the-art} hallucination detection capabilities.

Our key contributions are

\begin{enumerate}
    \item \textit{RealHall}: a suite of \textbf{four difficult, realistic benchmark datasets} for evaluating hallucination detection methods.
    \begin{itemize}
        \item We developed \textit{RealHall} by performing an extensive, careful review of academic papers and benchmarks on hallucination detection.
        \item We discovered that many of the datasets and benchmarks used to evaluate hallucination detection metrics in past work are \textbf{nearly irrelevant to practical users of today's LLMs}.
        \item LLMs have become much more powerful in just the past few years, and they're being deployed for a diverse range of difficult use cases. Evaluation benchmarks for LLMs have not caught up with this rapid progress.
        \item We developed \textit{RealHall} to close the gap between real LLM use and evaluation. \textit{RealHall} gives us confidence that our experiment results will generalize to real use cases.
    \end{itemize}

    \item \textit{ChainPoll}: a novel approach to hallucination detection that is \textbf{substantially more accurate} than any metric we've encountered in the academic literature.
    \begin{itemize}
        \item \textbf{\textit{ChainPoll} dramatically out-performs a range of published alternatives} -- including \emph{SelfCheckGPT} \citep{manakul2023selfcheckgpt}, \emph{GPTScore} \citep{fu2023gptscore}, \emph{G-Eval} \citep{liu2023geval}, and \emph{TRUE} \citep{honovich2022true} -- in a head-to-head comparison on \textit{RealHall}.
        \item \textit{ChainPoll} is also \textbf{faster and more cost-effective} than most of the metrics listed above.
        \item Though much of the research literature concentrates on the the easier case of \textit{closed-domain} hallucination detection, \textbf{we show that \textit{ChainPoll} is equally strong when detecting \emph{either} open-domain or closed-domain hallucinations.}
        \begin{itemize}
            \item We develop versions of \emph{ChainPoll} specialized to each of these cases: \emph{ChainPoll-Correctness} for open-domain and \emph{ChainPoll-Adherence} for closed-domain. 
            \item \textbf{The Correctness and Context Adherence metrics in the Galileo console are powered by \emph{ChainPoll-Correctness} and \emph{ChainPoll-Adherence}}, respectively.
        \end{itemize}
    \end{itemize}
\end{enumerate}

\vspace{2pt}
\begin{table}[H]
    \begin{tabularx}{\textwidth}{bb}
        \hline
         \textbf{Metric} & \textbf{Aggregate AUROC}  \\
         \hline
\textbf{\textit{ChainPoll}} & \mybarbold{0.781}{0.561}{6}{154}{243} \\ 
\hline

SelfCheck-Bertscore & \mybar{0.673}{0.345}{197}{200}{56} \\ 
\hline

SelfCheck-NGram & \mybar{0.644}{0.288}{248}{213}{6} \\ 
\hline

G-Eval & \mybar{0.579}{0.158}{255}{150}{0} \\ 
\hline

Max pseudo-entropy & \mybar{0.550}{0.101}{255}{120}{0} \\ 
\hline

GPTScore & \mybar{0.524}{0.047}{255}{93}{0} \\ 
\hline

Random Guessing & \mybar{0.500}{0.000}{255}{69}{0} \\ 
\hline
\hline
    \end{tabularx}
    \caption{Hallucination detection performance on \emph{RealHall}, averaged across datasets.}
    \label{tab:results-topline}
\end{table}

\subsection{Organization}

The rest of the paper is organized as follows.

\begin{itemize}
    \item \textbf{Section \ref{problemstatement}} describes the problem we're solving and outlines our basic methodology.
    \item \textbf{Section \ref{benchmarksection}} describes our critical review of existing datasets and benchmarks, and introduces our benchmark suite \emph{RealHall}.
    \item \textbf{Section \ref{metricssection}} surveys the metrics we evaluated in our research, including our best-performing metric \emph{ChainPoll}.
    \item \textbf{Section \ref{resultssection}} contains our experimental results.
    \item \textbf{Section \ref{otherpapers}} reviews past work on hallucination detection metrics.

\end{itemize}

\section{Problem statement}\label{problemstatement}

We are interested in the following setting:

\begin{itemize}
    \item We have a dataset of text inputs to a \textbf{state-of-the-art generative LLM} (large language model).
    \item We send the inputs to the LLM, and get back text \textbf{completions}, one for each input.
    \item We want to determine which of the completions, if any, contain hallucinations.
    \item We are interested in detecting both types of hallucination delineated in prior work \citep{openai2023gpt4}: \emph{open-domain} and \emph{closed-domain}.
    \begin{itemize}
        \item \emph{Open-domain hallucinations} are false claims about the world made by the LLM.
        \item \emph{Closed-domain hallucinations} involve the model straying from the context of a specific reference text, such as a document to summarize.
    \end{itemize}
    \item We want to identify hallucinations using one or more metric(s) that can be automatically computed, efficiently and at low cost.
    \item In some cases, we may be querying the model through an API like OpenAI’s. This limits the available information about it.
    \begin{itemize}
        \item We cannot use metrics that require access to model weights, activations, embeddings, or other information that would not be available through an API.
    \end{itemize}
    \item Our metric should work well across a \emph{diverse} range of tasks that are 
    \begin{itemize}
        \item \emph{challenging} enough to elicit frequent hallucinations, and 
        \item \emph{relevant}, in the sense of that they measure LLM capabilities that underlie practical use cases
    \end{itemize}
\end{itemize}

There are some important differences between the way we've framed the problem above, and the way it is typically framed in the academic literature:

\begin{enumerate}
    \item We have more exacting standards for quality. We require that our metrics perform well across a range of different tasks -- not just one or two -- and we require that these tasks are both \emph{challenging} and \emph{relevant}.
    \item Academic hallucination benchmarks are typically built around responses from older models that are much weaker than modern LLMs (e.g. \citep{fabbri2021summeval, wang2020asking, dziri2022evaluating}). These models often hallucinate in extreme ways that are relatively easy to detect. We seek metrics that can detect the subtler hallucinations produced by modern LLMs.
    \item Although much of the academic literature (e.g. \citep{liu2023geval, honovich2022true, zhong2022unified}) focuses solely closed-domain hallucination, we also aim to detect open-domain hallucinations.
    \item We aim to create practical methods that can be deployed as part of a product while maintaining a fast, fluid user experience. Some metrics in the academic literature can take hours to compute over a full dataset, even with a very powerful GPU; we require our metrics to be much more efficient than this.
\end{enumerate}

\subsection{Approach}

We treat hallucination detection as a binary classification problem. For our purposes, a \emph{metric} for hallucination detection is a binary classifier which outputs a scalar score\footnote{We do not assume that these scalar scores are probabilities, merely that they are ordered, with larger values indicating a higher likelihood of the positive class (hallucination).}.

To assess the performance of our metrics, we constructed four \emph{benchmark datasets}, which we collectively call \emph{RealHall} (Section \ref{benchmarksection}).

By a \emph{benchmark dataset}, we mean a list of prompts, completions and ground-truth boolean labels indicating whether each completion contained hallucination(s).

We use \emph{RealHall} to evaluate a variety of metrics (Section \ref{metricssection}), covering a range of different approaches proposed in prior work as well as our own novel metrics like \emph{ChainPoll}. 

\section{\textit{RealHall}}\label{benchmarksection}

\textit{RealHall} is a new benchmark suite for evaluating hallucination detection metrics, built on the guiding principles of \emph{Challenge}, \emph{Realism}, and \emph{Task Diversity} (Table \ref{tab:criteria}).

To build \textit{RealHall}, we conducted an extensive survey of available datasets,
applying the rubric given in Table \ref{tab:criteria}.

\begin{table}[H]
     \begin{tabularx}{\textwidth}{sb}
    \hline
           \textbf{Criterion} & \textbf{Description} \\ \hline

          Challenge & The LLM is asked to perform a task that is challenging, even for today's state-of-the-art LLMs. \\ \hline

           Realism & The task is relevant to practical use cases. \\ \hline

           Task Diversity & Taken as a whole, the benchmark suite should assess a wide range of different LLM capabilities. \\ \hline

    \end{tabularx}
    \caption{
    Criteria we applied when reviewing datasets.
    }
    \label{tab:criteria}
\end{table}

Most of the datasets we reviewed did not meet our bar for \emph{Challenge}, \emph{Realism}, and/or \emph{Task Diversity}.  Most notably, we found that \textbf{many benchmarks used in prior work on hallucination detection were deficient in one of more of these aspects}.

This observation is alarming, as it suggests that published evaluation results in this field do not provide reliable guidance about which metrics will perform well in real, practical use.

We created \emph{RealHall} to remedy this defect in past evaluations. \emph{RealHall} contains four datasets carefully selected for \emph{Challenge}, \emph{Realism}, and \emph{Task Diversity}.

For details on the process of constructing \emph{RealHall}, see Appendix \ref{datadetails}.

\subsection{\emph{RealHall} datasets} \label{dataused}

\textit{RealHall} contains four datasets, divided into two groups of two: \emph{RealHall Closed} and \emph{RealHall Open}.

\begin{itemize}
    \item \emph{RealHall Closed} evaluates how well a metric can detect \emph{closed-domain hallucinations}: inconsistency between the generated text and a reference text provided in the prompt.
    \item \emph{RealHall Open} evaluates how well a metric can detect \emph{open-domain hallucinations}: false claims about the real world.
\end{itemize}

\subsubsection{\emph{RealHall Closed}}

\emph{RealHall Closed} contains the datasets \textbf{COVID-QA with retrieval} and \textbf{DROP}.

\begin{itemize}
    \item \textbf{COVID-QA with retrieval}.
    \begin{itemize}
        \item COVID-QA \citep{moller-etal-2020-covid} is a dataset containing question-answer pairs about Covid-19, constructed by biomedical experiments. 
        \item We construct a RAG-like dataset from COVID-QA, following the approach in \citep{siriwardhana2022improving}. We build a vector store over the 250k-passage reference corpus from \citep{siriwardhana2022improving}, using OpenAI API embeddings. During inference, we and retrieve the top $k=4$ passages and present them alongside the question. (We use these retrieved documents instead of the original reference documents packaged with COVID-QA.\footnote{We retrieved documents from a vector store, rather than using the documents provided with the dataset, to mimic the RAG use case as closely as possible.})
        \item COVID-QA with retrieval is a realistic test of a metric's ability to detect closed-domain hallucinations in \textbf{Retrieval Augmented Generation (RAG)} use cases. It is moderately challenging for SOTA LLMs.
    \end{itemize}
    \item \textbf{DROP}.
    \begin{itemize}
        \item DROP \citep{dua2019drop} is an open-book QA dataset containing questions that require discrete reasoning over multiple facts mentioned in a passage (for example, locating two numbers in the passage, then subtracting one from the other).
        \item DROP is challenging for SOTA LLMs.  We included it in \emph{RealHall} alongside COVID-QA because it assesses a distinct, and more challenging, notion of consistency with the provided documents\footnote{Evaluating consistency for DROP requires discrete reasoning, for the same reason that \emph{answering} DROP questions requires discrete reasoning.}.
    \end{itemize}
\end{itemize}

\subsubsection{\emph{RealHall Open}}

\emph{RealHall Open} contains the datasets \textbf{Open Assistant prompts} and \textbf{TriviaQA}.

\begin{itemize}
    \item \textbf{Open Assistant prompts}.
    \begin{itemize}
        \item The Open Assistant dataset \citep{köpf2023openassistant} contains dialog trees solicited as training data for a ChatGPT-like assistant.
        \item We used only the initial prompts, i.e. the first turn of each dialog tree.
        \item We judged these prompts to be a good test bed for eliciting open-domain hallucinations: they cover a diverse range of tasks and subject matter, they are often challenging even for SOTA LLMs, and they are more representative of the way LLMs are prompted in practice than the prompts found in large instruction-tuning datasets.
    \end{itemize}
    \item \textbf{TriviaQA}.
    \begin{itemize}
        \item TriviaQA \citep{joshi2017triviaqa} contains question-answer pairs written by trivia enthusiasts.
        \item TriviaQA was originally proposed as a reading comprehension benchmark, with reference documents provided to the model. However, recent works (e.g. \citep{touvron2023llama}) tend to use the questions alone, without the documents, to probe LLMs for factual knowledge. 
        \item We follow this line of work and present TriviaQA questions alone. TriviaQA in this format is challenging for SOTA LLMs, and we judged it to be a useful supplement to the signal we get from benchmarking metrics against Open Assistant prompts, focusing in on the LLM's ability to faithfully recall declarative knowledge.
    \end{itemize}
\end{itemize}

\section{Metrics}\label{metricssection}

\subsection{Metrics evaluated}

We benchmarked \emph{ChainPoll} against a wide range of other metrics from the literature on LLM hallucinations:

\begin{itemize}
    \item \emph{ChainPoll} without detailed CoT.
    \begin{itemize}
        \item This metric combines the same aggregation method as \emph{ChainPoll} (Section \ref{askagg}) with a ``vanilla'' chain-of-thought prompt, while \emph{ChainPoll} uses a more carefully engineered prompt we call ``detailed CoT.''
    \end{itemize}
    \item G-Eval-3.5\footnote{Referred to as simply ``G-Eval'' below for simplicity.}
    \item GPTScore
    \item SelfCheck-BertScore
    \item SelfCheck-NGram
    \item TRUE (closed-domain only)
\end{itemize}

See Section \ref{otherpapers} for descriptions of these metrics, and Table \ref{tab:other_metrics_table} for details on why we excluded some metrics from our evaluations.

We also include a simple probability-based baseline, following prior work (e.g. \citep{manakul2023selfcheckgpt}).

For our probability-based baseline, we use a metric we call \emph{pseudo-entropy}: an approximation to the Shannon entropy, adapted to settings like the OpenAI API in which only a subset of probabilities are available. We found that pseudo-entropy performed best across a range of probability-based metrics we investigated.

For a full description of pseudo-entropy, see Appendix \ref{pseudoentropy}.

\subsubsection{BLEU, ROUGE, METEOR and similar metrics}

We chose not to benchmark any metrics that compare the LLM's completion with a \emph{ground-truth response}. This category includes BLEU, ROUGE, and METEOR.

Metrics that require a ground-truth response cannot adequately serve the full range of hallucination-detection needs that LLM users face. While ground-truth responses may be available for some users in some parts of the LLM workflow, they are \emph{not} reliably available in key scenarios like

\begin{itemize}
    \item \emph{Monitoring}: an LLM application in production must respond to arbitrary user input. It's impossible to prepare a ground-truth response in advance for every possible input the user could send to the system.
    \item \emph{Rapid experimentation}: it's often unclear what LLMs are and aren't capable of, and users may want to rapidly ``try out'' many different hypothetical tasks for the LLM. Producing prompts for a novel task is much faster and easier than producing ground-truth responses, especially in creatively defined tasks where it's not clear at the outset what the correct response \emph{should} be.
\end{itemize}

\subsection{Defining \emph{ChainPoll}} \label{askagg}

Across all datasets, our best-performing metrics use the approach we call \textit{ChainPoll}.

To compute these metrics, we take the following steps:

\begin{enumerate}
    \item Ask \texttt{gpt-3.5-turbo} whether the completion contained hallucination(s), using a detailed and carefully engineered prompt.
    \item Run step 1 multiple times, typically 5. (We use batch inference here for its speed and cost advantages.)
    \item Divide the number of "yes" answers from step 2 by the total number of answers to produce a score between 0 and 1.
\end{enumerate}

Among metrics previously proposed in the literature, \emph{ChainPoll} is perhaps closest to G-Eval \citep{liu2023geval}.

However, we find that \emph{ChainPoll} dramatically outperforms G-Eval across the entirety of \emph{RealHall}. We attribute this to a number of key differences between \emph{ChainPoll} and G-Eval:

\begin{itemize}
    \item We put considerable effort into prompt engineering. In particular, we phrase our chain-of-thought prompt in a way that reliably elicits a very specific and systematic explanation from the model, an prompting approach we call ``detailed CoT.''
    \begin{itemize}
        \item By contrast, the prompts used in \citep{liu2023geval} either did not use chain-of-thought, or asked for the answer before the chain-of-thought explanation, which prevents the answer from leveraging the reasoning in the explanation.
        \end{itemize}
    \item We request boolean judgments, rather than numeric scores. In early experiments on this distinction, we observed that boolean judgments work better than scores, even when eliciting only a single completion.
    \item We use \texttt{gpt-3.5-turbo}, while \citep{liu2023geval} used either \texttt{text-davinci-003} or \texttt{gpt-4}\footnote{While GPT-4 performs very well at hallucination detection, we consider it too expensive for routine use in production. Our goal is here to offer high quality hallucination detection without the expense of GPT-4. (OpenAI's pricing is such that aggregating over multiple \texttt{gpt-3.5-turbo} completions is still much cheaper than generating a single completion with GPT-4.)}.
\end{itemize}

\subsubsection{\emph{ChainPoll-Correctness} and \emph{ChainPoll-Adherence}}

Depending on the situation, a user may want to detect open-domain hallucinations, closed-domain hallucinations, or both.

We define a \emph{ChainPoll}-based metric for each of these cases.

\begin{itemize}
    \item \emph{ChainPoll-Correctness} uses \emph{ChainPoll} to detect open-domain hallucination.
    \item \emph{ChainPoll-Adherence} uses \emph{ChainPoll} to detect open-domain hallucination.
\end{itemize}

The two metrics differ only in the prompt format used when prompting \texttt{gpt-3.5-turbo}. In \emph{ChainPoll-Correctness}, the prompt format asks the model to look for open-domain hallucinations; in \emph{ChainPoll-Adherence}, the prompt format asks the model to look for closed-domain hallucinations.

\subsubsection{Re-using chains of thought for explainability}

In the \textit{ChainPoll} approach, the LLM is asked to judge whether or not the original completion contained hallucination, justifying its answer with a chain-of-thought (CoT) explanation. While we use CoT primarily as a means to improve the quality of the model's final judgements, it also opens up interesting avenues for explaining that judgment to an end user.

That is, we can re-use the CoT text generated by the model as a \emph{justification} for the judgment that the completion did, or did not, contain hallucination(s)\footnote{An interesting line of recent work has called into question whether model-generated chains of thought faithfully reflect the model's actual reasoning process \citep{lanham2023measuring}. From our perspective, this does not reduce the value of model-generated chains of thought as user-facing \emph{justifications} for model-generated judgments.\\ For example, suppose the model tells us that a particular completion was hallucinated, and it provides a CoT consistent with this judgment. \emph{Whether or not} the CoT reflects how the model actually arrived at its judgment, it is nonetheless an argument that \emph{could} be made in favor of that judgment. If the argument is correct, then it has value for the end user, even if the model arrived at its (correct) judgment through some other reasoning process.}.

We find that the chains of thought written by \texttt{gpt-3.5-turbo} are coherent and frequently convincing. An example is given below, explaining why a COVID-QA completion was hallucinated:

\begin{quote}

\texttt{The response claims that the study described is a descriptive study of severe hospitalized cases of laboratory-confirmed influenza during five epidemic seasons (2010–2015).}

\texttt{To verify whether this claim is supported by the documents, we need to check if any of the documents mention a study or research that matches the description.}

\texttt{The first document mentions the need to deepen knowledge of the effect of aging, chronic diseases, and immune response in order to implement preventive measures. This document does not directly state that it describes a study of severe hospitalized cases of laboratory-confirmed influenza.}

\texttt{The second document does not provide any relevant information to support or refute the claim.}

\texttt{The third and fourth documents both mention a study design and protocol for investigating social network patterns and respiratory disease transmission and control. These documents do not mention severe cases of laboratory-confirmed influenza.}

\texttt{Based on the analysis of the documents, none of them provide direct support for the claim that the study described is a descriptive study of severe hospitalized cases of laboratory-confirmed influenza during five epidemic seasons (2010–2015).}

\texttt{Therefore, the response is not supported by the documents.}

\end{quote}

\section{Results}\label{resultssection}

We present the results of our evaluations here, demonstrating that \textbf{\emph{ChainPoll} is a new state-of-the-art.} 

Across a diverse range of benchmark tasks, the \emph{ChainPoll} outperforms all other methods -- in most cases, by a huge margin.

Taking efficiency into account \emph{ChainPoll}'s lead is even larger. It outperforms the next-best method, SelfCheck-BertScore, while using only 1/4 as much LLM inference, and without using an additional model like BERT.

Unlike all other methods considered here, \emph{ChainPoll} also provides human-readable verbal \emph{justifications} for the judgments it makes, via the chain-of-thought text produced during inference.

\subsection{AUROC scores}

\vspace{2pt}
\begin{table}[H]
    \begin{tabularx}{\textwidth}{bb}
        \hline
         \textbf{Metric} & \textbf{Average AUROC}  \\
         \hline

\emph{ChainPoll-Correctness} & \mybarbold{0.772}{0.544}{6}{154}{243} \\ 
\hline

SelfCheck-Bertscore & \mybar{0.670}{0.339}{193}{199}{60} \\ 
\hline

SelfCheck-NGram & \mybar{0.636}{0.272}{254}{214}{0} \\ 
\hline

G-Eval & \mybar{0.574}{0.148}{255}{148}{0} \\ 
\hline

Max pseudo-entropy & \mybar{0.565}{0.131}{255}{138}{0} \\ 
\hline

GPTScore & \mybar{0.489}{-0.021}{255}{69}{0} \\ 
\hline

Random Guessing & \mybar{0.500}{0.000}{255}{69}{0} \\ 
\hline

    \end{tabularx}
    \caption{Open-domain hallucination detection performance on \emph{RealHall Open}, averaged across datasets.}
    \label{tab:results-open}
\end{table}

\vspace{2pt}
\begin{table}[H]
    \begin{tabularx}{\textwidth}{bb}
        \hline
         \textbf{Metric} & \textbf{Average AUROC}  \\
         \hline
         \emph{ChainPoll-Adherence} & \mybarbold{0.789}{0.579}{6}{154}{243} \\ 
\hline

SelfCheck-Bertscore & \mybar{0.675}{0.351}{201}{201}{52} \\ 
\hline

SelfCheck-NGram & \mybar{0.652}{0.303}{244}{212}{10} \\ 
\hline

TRUE & \mybar{0.593}{0.186}{255}{162}{0} \\ 
\hline

G-Eval & \mybar{0.584}{0.167}{255}{153}{0} \\ 
\hline

Max pseudo-entropy & \mybar{0.535}{0.071}{255}{104}{0} \\ 
\hline

GPTScore & \mybar{0.558}{0.115}{255}{127}{0} \\ 
\hline

Random Guessing & \mybar{0.500}{0.000}{255}{69}{0} \\ 
\hline
    \end{tabularx}
    \caption{Closed-domain hallucination detection performance on \emph{RealHall Closed}, averaged across datasets.}
    \label{tab:results-closed}
\end{table}

Per-dataset AUROC scores are provided in Appendix \ref{detailedresults}.

\subsection{Plots}

\begin{figure}[H]
    \centering
    \includegraphics[width=0.8\textwidth]{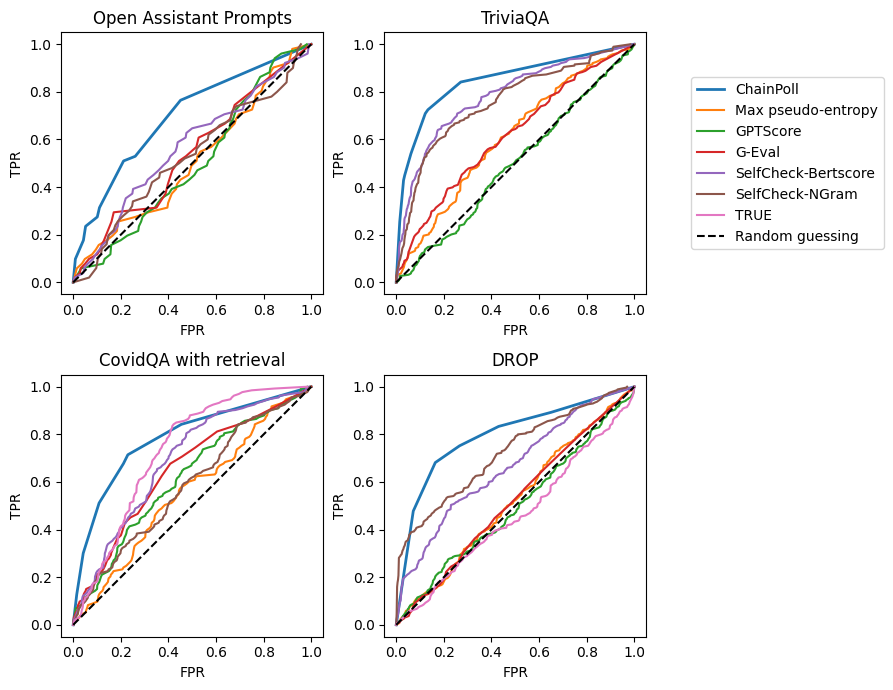}
    \caption{ROC curves for hallucination detection across \emph{RealHall} datasets. The ChainPoll curves are \emph{ChainPoll-Correctness} in the top row, and \emph{Chainpoll-Adherence} in the bottom row.}
    \label{fig:combined_roc}
\end{figure}

\begin{figure}[H]
    \centering
    \includegraphics[width=0.8\textwidth]{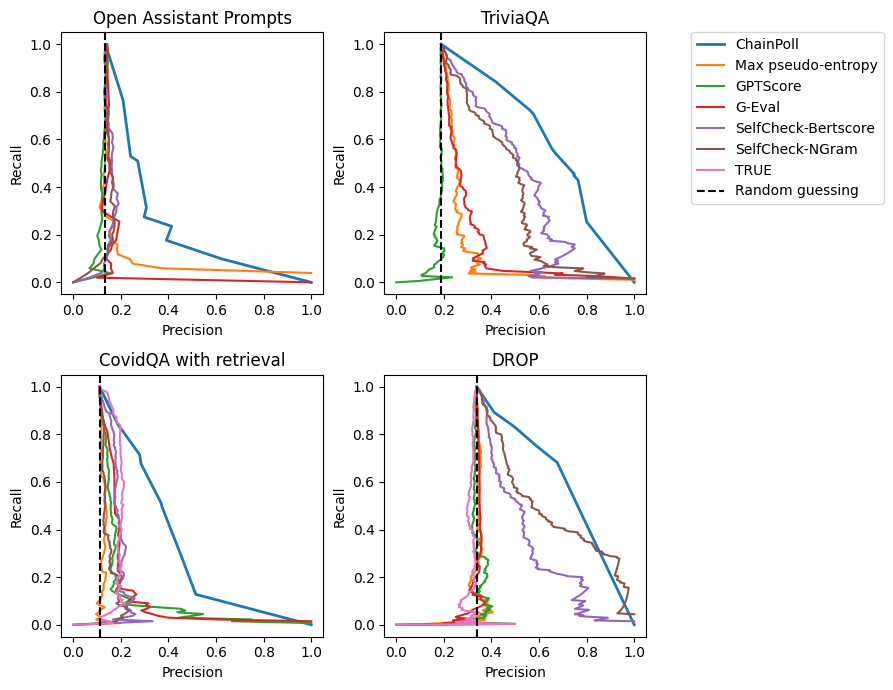}
    \caption{Precision-recall curves for hallucination detection across \emph{RealHall} datasets. The ChainPoll curves are \emph{ChainPoll-Correctness} in the top row, and \emph{Chainpoll-Adherence} in the bottom row.}
    \label{fig:combined_pr}
\end{figure}




\section{Related work}\label{otherpapers}

The field of LLM hallucination detection is relatively new, as LLMs themselves are relatively new.

Rather than giving a complete historical review of this field, we will cover the specific metrics that we deemed most promising when reviewing the literature.

Table \ref{tab:other_metrics_table} provides a summary view of these metrics, comparing and contrasting them to our best-performing metric, \emph{ChainPoll}.

\begin{table}[H]

     \begin{tabularx}{\textwidth}{bbbsbsss}
    \hline
           Metric & Description & Tested against$*$ & Type$^a$ & Cost/ex$^b$ & Batch$^c$ & GPU-free$^d$ & Quality vs ours$^e$ \\ \hline
           
           \emph{ChainPoll} (ours) & Prompting + aggregation & \emph{RealHall} & OC & 5 & \checkmark & \checkmark & 100\%\\ \hline

           \emph{SelfCheck-BertScore} \citep{manakul2023selfcheckgpt} & Sentence-level rerun checking & Wikipedia articles & O & >20 & \checkmark  & & 63\% \\ \hline

           \emph{SelfCheck-NGram} \citep{manakul2023selfcheckgpt} & Sentence-level rerun checking & Wikipedia articles & O & 20 & \checkmark & \checkmark  & 52\% \\ \hline

           \emph{TRUE NLI} \citep{honovich2022true} & NLI & Various (closed-domain)  & C & $\le 1$ & \checkmark & & 33\%$^h$\\ \hline

           \emph{G-Eval-3.5}$^f$ \citep{liu2023geval} & Prompting + aggregation & SummEval, TopicalChat & C & 20 & \checkmark & \checkmark & 29\%\\ \hline

           \emph{GPTScore} \citep{fu2023gptscore} & Prompting + perplexity & Various (closed-domain) & C & $\le 1$ & \checkmark & \checkmark & 9\%\\ \hline

           \emph{SelfCheck-MQAG} \citep{manakul2023selfcheckgpt} & Sentence-level rerun checking & Wikipedia articles & O & >20 & &   & N/A$^g$ \\ \hline

          \emph{ChatProtect} \citep{mündler2023selfcontradictory} & Sentence-level rerun checking & Wikipedia articles & O & 2/sentence & &  \checkmark  & N/A$^g$ \\ \hline

    \end{tabularx}
    
    \caption{How our metrics compare to others in the literature. 
$^*$ This column lists the evaluation data used to test the quality of the metric in the original publication introducing it. In this paper, we independently evaluate all metrics on \emph{RealHall}. 
$^a$ O = open-domain, C = closed-domain, OC = both open-domain and closed-domain 
$^b$ An estimate of how compute-intensive the metric is, in units of \emph{additional LLM-generated responses} required during evaluation of a single response. $>N$ denotes a metric which generates $N$ completions and then does additional computation with a neural model. 
$^c$ Whether the computations noted in the \emph{Cost/ex} column can be computed in parallel. In practice, batch metrics are much faster than sequential metrics, even if they require more computation per example. 
$^d$ Whether the metric can be served without the added expense of a dedicated GPU. 
$^e$ Average (AUROC score minus 0.5) over datasets in \emph{RealHall}, as a fraction of \emph{ChainPoll}'s performance. We subtract 0.5 to normalize scores because an AUROC score of 0.5 corresponds to random guessing. 
$^f$ We do not benchmark G-Eval-4, as it does not meet our bar for cost (it is 20x as expensive as the already-expensive GPT-4). 
$^g$ We did not benchmark these metrics, as their high compute intensity and sequential nature do not meet our bar for efficiency. 
$^h$ Closed-domain tasks only, as TRUE NLI cannot be applied in open-domain tasks.}
\label{tab:other_metrics_table}

\end{table}

\subsection{SelfCheckGPT}

SelfCheckGPT \citep{manakul2023selfcheckgpt} proposed an approach based on checking the self-consistency between an LLM response and a large number of additional responses, sampled from the same LLM using the same prompt. For their main experiments, the authors used 20 additional responses per evaluated response.

The authors introduced three metrics using this approach, which differ in the way they compute agreement between responses.

\begin{itemize}
    \item \emph{SelfCheck-BertScore} computes agreement using BertScore \citep{zhang2020bertscore}.
    \item \emph{SelfCheck-NGram} computes agreement by fitting a simple unigram\footnote{The authors experimented with different gram lengths, finding that unigrams worked best.} language model and using its probabilities on the original response.
    \item \emph{SelfCheck-MQAG} computes agreement using a MQAG \citep{manakul2023mqag}, a complex question-answering pipeline using four fine-tuned neural models. The pipeline generates multiple-choice questions based on the original response, then tries to answer them using only the additional responses.
\end{itemize}

Notably, the SelfCheckGPT metrics were proposed as \emph{sentence-level} metrics, requiring computation of agreement scores separately for each sentence in the response -- which can be computationally expensive for long responses. (This expense combines with the expense of generating a potentially large number of additional responses.)

To compute response-level aggregates, the scores for sentences are averaged.

The SelfCheckGPT metrics were evaluated on a dataset of 238 prompts written by the authors, all of which ask the model to write a Wikipedia page for a specific person. We critically assess this dataset in Section \ref{datarej}.

\subsection{G-Eval}

G-Eval \citep{liu2023geval} propose an approach that evaluates an LLM response by asking an LLM\footnote{Either a different LLM, or the same one.} to rate the response on a 1-5 scale, with provided guidelines.

The LLM's probabilities of outputting the tokens \texttt{' 1', ' 2',  etc.} are used to produce a weighted-average rating. When probabilities are not available, the authors sample from the LLM 20 times and average the ratings over this sample.

G-Eval was evaluated on two closed-domain datasets, SummEval and TopicalChat. We critically assess these datasets in Section \ref{datarej}, with further analysis in Appendix \ref{summeval}.

\subsection{GPTScore}

GPTScore \citep{fu2023gptscore} uses a simple method which evaluates an LLM response by prepending a instruction (e.g. ``write a factually consistent summary'') to the prompt and response, then evaluating the \emph{perplexity} of the response using an LLM\footnote{As in G-Eval, this may be a different LLM, or the same one.}.

GPTScore was evaluated on a large number of closed-domain datasets, such as SummEval.

We reproduce GPTScore's strong performance on SummEval (Appendix \ref{summeval}), yet we find that it performs very poorly on \emph{RealHall}.

We hypothesize that this discrepancy results from the following:

\begin{itemize}
    \item We find that the prefix adds little value: we can achieve nearly identical performance on SummEval by simply computing perplexity on the original response.
    \item We analyze the strong performance of perplexity on SummEval, finding that it results from a mismatch between the (weaker) models used to generate responses in SummEval and the (stronger) models used to compute perplexity.
    \item When a strong, modern LLM is used \emph{both} to generate responses and to compute perplexity, the strong performance of perplexity (and thus GPTScore) disappears.
\end{itemize}

\subsection{TRUE}

TRUE \citep{honovich2022true} builds a benchmark suite of 11 closed-domain datasets, covering similar ground to the evaluation datasets used in GPTScore \citep{honovich2022true}.

The authors compare a number of different metrics on these datasets. Their best-performing metric used probabilities from a T5-XXL model finetuned for natural language inference (NLI). 

In conjunction with the paper, the authors released the weights of a model similar to this one, though trained on a different data mixture. We compute scores using the released model, and refer to this metric as \emph{TRUE NLI}.

\subsection{ChatProtect}

ChatProtect \citep{mündler2023selfcontradictory} proposes an approach similar to SelfCheckGPT \citep{manakul2023selfcheckgpt}.

Like SelfCheckGPT, ChatProtect works on the sentence level, and uses self-consistency between multiple responses to detect hallucinations.

Whereas SelfCheckGPT generates alternatives at the \emph{response}, ChatProtect generates a separate alternative version of each \emph{sentence} in context, and checks each one for consistency against the original sentence.

\section{Conclusion}

Hallucinations are possibly the single largest impediment to widespread practical use of LLMs. This fact means there is a pressing need to identify ways of automatically discovering hallucinations in LLM outputs.

We have developed a benchmark suite, \emph{RealHall}, for evaluating these hallucination detection metrics. Notably, we find that many tasks and datasets used in past work have minimal relevance to practical use of SOTA LLMs. Our benchmark suite focuses in on four practically relevant tasks on which even today's powerful LLMs hallucinate with alarming frequency.

We use our benchmark suite to evaluate a variety of metrics for open-domain and closed-domain hallucination detection -- including a new metric, \emph{ChainPoll}, which outperforms all other metrics considered, while being efficient to compute and inherently explainable.

\bibliographystyle{unsrtnat}
\bibliography{main}

\appendix

\section{Appendices}

\subsection{Dataset selection process}\label{datadetails}

Table \ref{tab:datasets_table} lists all the datasets we reviewed during the development of \emph{RealHall}.

\ref{datarej} covers the datasets we reviewed but did not include in \emph{RealHall}.

\ref{dataannot} describes our process for assigning ground-truth labels to the benchmark data.



\begin{table}[H]
     \begin{tabularx}{\textwidth}{bbsbs}
    \hline
           Dataset & Description & Type* & Notes & Included in \emph{RealHall} \\ \hline
           
           Open Assistant Prompts \citep{köpf2023openassistant} & Prompts for an LLM assistant & O &   & \checkmark  \\ \hline

           TriviaQA \citep{joshi2017triviaqa} & General knowledge questions & O$\dagger$ &  & \checkmark  \\ \hline

           Self-Instruct Human Eval \citep{wang2023selfinstruct} & Prompts for an LLM assistant & O & Less challenging than Open Assistant &  \\ \hline

          Super-NaturalInstructions \citep{wang2022supernaturalinstructions} & Instruction tuning data & O, C & Not reflective of practical LLM use &  \\ \hline

          FLAN \citep{longpre2023flan} & Instruction tuning data & O, C & Not reflective of practical LLM use &  \\ \hline

           SelfCheckGPT Wikibio \citep{manakul2023selfcheckgpt} & Prompts of the form ``write a Wikipedia article about X'' & O & Narrow task, memorization concerns & \\ \hline 

            HaluEval \citep{li2023halueval} & Prompts and completions with synthetic hallucination & O, C & Not representative of naturally arising hallucination & \\ \hline 
            
           COVID-QA \citep{moller-etal-2020-covid} with retrieval \citep{siriwardhana2022improving} & Covid-19 knowledge questions & C &  & \checkmark  \\ \hline

           DROP \citep{dua2019drop} & Discrete reasoning questions & C &  & \checkmark  \\ \hline

            NarrativeQA \citep{kočiský2017narrativeqa} & Reading comprehension questions & C & Noisy labels, easy for SOTA LLMs & \\ \hline 

           SummEval \citep{fabbri2021summeval} & News summarization prompts and completions & C & See Section \ref{datasumm} & \\ \hline 

           TL;DR \citep{volske-etal-2017-tl} & Reddit summarization prompts and completions & C & See Section \ref{datasumm} & \\ \hline 

            ArXiV Summarization \citep{cohan-etal-2018-discourse} & Scientific paper summarization prompts & C & See Section \ref{datasumm} & \\ \hline 

            MediaSum \citep{zhu2021mediasum} & Interview summarization prompts & C & See Section \ref{datasumm} & \\ \hline 

           BEGIN \citep{dziri2022evaluating} & Knowledge-conditioned dialogue & C & See Section \ref{otherrejected} & \\ \hline 

    \end{tabularx}
    \caption{
    Datasets reviewed during benchmark construction.
    * O = open-domain, C = closed-domain
    $\dagger$ We present TriviaQA questions on their own, without reference documents.
    }
    \label{tab:datasets_table}
\end{table}

\subsubsection{Rejected datasets} \label{datarej}

We distilled \emph{RealHall} from a long list of candidate datasets by applying the rubric given in Table \ref{criteria}.

Most of the datasets we considered did not meet this bar. Here, we detail the reasons behind our choice not to include various datasets in \emph{RealHall}.

\paragraph{Instruction tuning datasets.} \label{datainst} Instruction tuning datasets, such as Super-NaturalInstructions \citep{wang2022supernaturalinstructions} and FLAN \citep{longpre2023flan}, were ruled out by our second criterion: \emph{The tasks and prompts should be reflective of real LLM use ``in the wild''}.

While these datasets work well as \emph{training} data for instruction-tuning an LLM, we concluded that they are not well representative of the way users \emph{interact} with instruction-tuned LLMs in practice.

To illustrate this claim, consider two selected instructions from \\ Super-NaturalInstructions: 

\begin{itemize}
    \item \emph{``Generate a question, given a collection of facts.''}
    \item \emph{``Given premise, initial context with ending, and new counterfactual ending, generate counterfactual context which supports the new story ending.''}
\end{itemize}

Highly artificial, structured tasks like this may be helpful if you want to train an LLM to understand instructions, but they bear little resemblance to the tasks and instructions seen during practical LLM use. 

\paragraph{Summarization datasets.} \label{datasumm} In the academic literature, the summarization datasets Summeval \citep{fabbri2021summeval} and QAGS \citep{wang2020asking} are often\footnote{For example, Summeval was used in this fashion in \citep{liu2023geval, honovich2022true, zhong2022unified, tam2022evaluating, chen2023evaluating}.} used to evaluate metrics for closed-domain hallucination detection. 

These datasets share the following key features:

\begin{enumerate}
    \item They consist of 
    \begin{itemize}
        \item \emph{documents} from a standard summarization dataset, e.g. CNN/DM, together with
        \item model-written \emph{summaries} of these documents, and
        \item human annotations assessing the quality of the summaries
    \end{itemize}
    \item The models used to generate for the model-written summaries are much \emph{weaker} than current SOTA LLMs.
\end{enumerate}

These datasets are easy to use as hallucination benchmarks, because they come packaged with human annotations (point 1).

However, because the summaries packaged with these datasets were produced by much weaker models than modern LLMs (point 2), these datasets do not reflect the distribution of hallucination-like behaviors observed with today's SOTA LLMs during practical use.

Our experiments show that modern SOTA LLMs hallucinate much less often in summarization tasks than the older models used in Summeval and QAGS.

For example, as assessed by GPT-4\footnote{See Section \ref{dataannot} for more on our use of GPT-4.},

\begin{itemize}
    \item 20\% of the summaries in Summeval contain hallucination(s)
    \item when ChatGPT\footnote{Unless otherwise noted, we use the terms ``ChatGPT'' and \texttt{gpt-3.5-turbo} interchangeably. Although matches common practice, we clarify it explicitly here because the ChatGPT product includes a GPT-4 option for paid subscription users.} is asked to summarize the same set of documents, only 3\% of the summaries contain hallucination(s)
\end{itemize}

We observed a similarly low rate of hallucination on many other summarization datasets, including TL;DR, ArXiv Summarization, and MediaSum.

We conclude that summarization is ``too easy'' for SOTA LLMs to make it a good benchmark for hallucination detection; while these models still hallucinate occasionally in summarization, they do it infrequently enough that a large amount of data would be necessary to distinguish signal from noise when making comparisons between metrics. Thus, we focus on other tasks that SOTA LLMs find more difficult.

Given the popularity of SummEval as a benchmark, we include SummEval evaluations in an Appendix (\ref{summeval}), though we do not include it in \emph{RealHall}.

\paragraph{Other rejected datasets.}\label{otherrejected}

This section covers our reasons for rejecting other datasets that do not fall under the broad themes covered in \ref{datainst} and \ref{datasumm}.

\begin{itemize}
    \item BEGIN, introduced in \citep{dziri2022evaluating}, contains model-written responses generated for three dialogue datasets (WOW, TopicalChat and CMU) from four models.
    \begin{itemize}
        \item We did not include this dataset because the models (e.g. GPT-2, CTRL) are much weaker than today's LLMs.
        \item We ran a small experiment benchmarking some metrics on the TopicalChat subset of BEGIN, and found similar trends to those observed in Appendix \ref{summeval}, e.g. the anomalously strong performance of perplexity.
    \end{itemize}
    \item SelfCheckGPT Wikibio, introduced in \citep{manakul2023selfcheckgpt}, contains model-written imitations of biographical Wikipedia papers, labeled for factual accuracy at the sentence level.
    \begin{itemize}
        \item We did not include this dataset because of its narrow scope, and because we noticed a number of verbatim-memorized Wikipedia pages among the \emph{non-}hallucinated examples.\footnote{Although memorized content repeated by LLMs is often factually accurate, it is atypical of LLM-generated content, and may provide misleading signals about which metrics work well in the general case. Memorized text can often be detected through its anomalously high model likelihood. Thus, when the hallucinated/not-hallucinated distinction is confounded by the memorized/not-memorized distinction, the hallucination detection performance of likelihood-based methods will be overestimated.}
    \end{itemize}
    \item HaluEval \citep{li2023halueval} contains several datasets. With the exception of the ``general'' set, each HaluEval set contains a mixture of (1) ``clean'' ChatGPT completions generated in the ordinary manner, and (2) synthetic hallucinations, i.e. hallucinated completions generated by explicitly asking ChatGPT to hallucinate.
    \begin{itemize}
        \item We did not include this dataset because we concluded that the synthetic hallucinations were not a representative proxy of ``real'' hallucinations produced naturally during ChatGPT inference.
        \item For example, on the HaluEval QA dataset, the synthetic-hallucination answers were typically much longer than the ``clean'' answers. (The average character lengths were 66 and 14, respectively; 90\% of the ``clean'' answers were under 25 characters, yet the same is true for 5\% of the  synthetic-hallucination answers.) It would be trivial to design a metric that could distinguish these two types of answers, but this ``success'' would not transfer to the case of real, naturally occurring hallucinations.
    \end{itemize}
    \item NarrativeQA \citep{kočiský2017narrativeqa} contains reading comprehension questions based on stories (books or film scripts). Following common practice, we used the summaries included in NarrativeQA as reference documents, rather than the much longer original texts.
    
    \begin{itemize}
        \item We did not include this dataset for a combination of reasons.
        \item During initial testing, we found that GPT-4 marked 8\% of ChatGPT-generated answers as hallucinated, while marking 12\% of \emph{ground truth} answers as hallucinated (!). Digging in further, we discovered that NarrativeQA questions are often ill-posed -- the answer is not available in the summary\footnote{The questions were written on the basis of the summaries alone, so using summaries rather than full texts does not account for this issue.}.
        \item Most of the hallucinations in the ChatGPT data occurred when the question was ill-posed. Ignoring these cases, ChatGPT almost never hallucinated on NarrativeQA. We conclude that this task is too easy to make a good benchmark for our purposes.
    \end{itemize}
\end{itemize}

\subsection{Model completions} \label{datageneration}

After selecting datasets for our benchmarks, we generated completions for each example in each benchmark using an LLM from the OpenAI API.

We used \texttt{gpt-3.5-turbo}, commonly known as ChatGPT, to generate completions in most of our experiments. In an early set of experiments on the Open Assistant Prompts dataset, we also experimented with \texttt{text-davinci-003} as the completion model.

When necessary, we wrote simple prompt formats (e.g. \texttt{'Answer the question, using the documents. \{question\} \{documents\}'}) to communicate the task to the model. 

\subsection{Data annotation} \label{dataannot}

We assigned a boolean ground-truth label to each (prompt, completion) pair: 1 if the completion contained any hallucination(s), 0 otherwise.

To produce these labels, we used a mixture of human annotation, GPT-4 annotation, and automatic rule-based scoring.

Our earliest experiments used human annotations on the Open Assistant dataset. Concurrently with this early work, we constructed a carefully engineered prompt for GPT-4 which asked it to determine whether a completion from another model contained hallucination(s).

We found that GPT-4 performed extremely well as an annotator, in the sense that it agreed very closely with the judgments of our human annotators. In fact, GPT-4 disagreed with our human annotators no more often than the human annotators disagreed with one another.

Encouraged by this result, we used GPT-4 as the sole annotator for several of the datasets considered here, specifically COVID-QA and TriviaQA. 

The DROP dataset is conventionally evaluated using a bag-of-words-based F1 score. We used this score to assign ground-truth labels for DROP, since we observed that it produced very similar results to GPT-4 while being faster to compute.\footnote{We used \texttt{lm-eval-harness} \citep{eval-harness} to compute DROP F1 scores. To convert these to boolean labels, we marked scores of 0 as hallucinated, and any score above 0 as not hallucinated. Thresholding at zero yielded better agreement with GPT-4 than other thresholds we tried.}

\subsection{Pseudo-entropy}\label{pseudoentropy}

The OpenAI API only provides probabilities for a subset of possible tokens at each position. The model's full vocabulary covers tens of thousands of tokens, but the API only provides probabilities for 5 or 6\footnote{The API returns the probability of the sampled token, as well as the probability of the top 5 most likely tokens. Thus it returns 5 probabilities if the sampled token is in the top 5, and 6 otherwise.}.

Let $N$ be the size of the full vocabulary, and $M$ be the number of tokens for which probability data is available through the API, where $M \ll N$.  Let $p_i$ be the probability of the $i$th token. Without loss of generality, suppose the tokens are ordered so that the $M$ tokens with API-supplied probabilities appear first.

The Shannon entropy of the distribution is 

\begin{equation}
    S = \sum_{i=1}^N p_i \log\left(p_i\right)
\end{equation}

but we cannot compute this exactly because $p_{M+1}, \ldots, p_N$ are unavailable.

PPL5, introduced in \citep{manakul2023selfcheckgpt}, makes the following approximation in this case. Let $\tilde{p}_i$ be the probability obtained by normalizing the top $M$ probabilities so they sum to one\footnote{Equivalently, by applying a softmax operation to the top $M$ log probabilities.}:

\begin{equation} \label{pi_eqn}
    \tilde{p}_i = - \frac{p_i}{\sum_{i=1}^M p_i}
\end{equation}

Then PPL5 is the Shannon entropy, computed with $\tilde{p}_i$ instead of $p_i$:

\begin{equation}
    \textrm{PPL5} = - \sum_{i=1}^N \tilde{p} \log\left(\tilde{p}\right)
\end{equation}

Consider the case in which most of the probability mass lies outside the top $M$ tokens. In this case, the true Shannon entropy will be large (all else being equal), since the distribution is spread out over many outcomes. However, the normalization in (\ref{pi_eqn}) removes all information about the amount of mass contained in the rest of the distribution, causing PPL5 to ignore this information and yielding an undesirably low estimate of the entropy.

To remedy this defect, we introduce a variant we call \emph{pseudo-entropy}:

\begin{equation}
    \textrm{Pseudo-entropy} = - \sum_{i=1}^N \tilde{p} \log\left(p\right)
\end{equation}

The difference lies in the use of $p$, rather than the normalized $\tilde{p}_i$, in the log-probability term.

When the distribution is spread out, the $M$ values of $p_i$ will be relatively low, and this fact will propagate through this term to yield a lower estimate of the entropy, as desired.

Galileo's \textbf{Uncertainty Score} is a transformed version of the pseudo-entropy. Specifically, we use a scaled and shifted expit transform to convert the pseudo-entropy into a probability, setting the scale and shift constants so that it is an unbiased predictor of ground-truth hallucination on our Open Assistant Prompts benchmark.

\subsubsection{Probability models}

Interestingly, we found that the performance of probability-based metrics is not strongly dependent on the model used to produce the token probabilities.

The OpenAI API does not return token probabilities for \texttt{gpt-3.5-turbo} at this time, so when it is the completion model, we must use a different model as the probability model.

In our experiments, we observed comparable performance across the probability models \texttt{text-curie-001}, \texttt{text-davinci-003}, and the recently introduced \texttt{davinci-002}, despite the different in size between the former and the latter two. The pseudo-entropy scores reported here were computed using \texttt{text-curie-001}.

\subsection{Evaluation details}

When computing metrics from past work, we used the code, models and prompts released by the original authors wherever possible.

In the case of G-Eval and GPTScore, the prompts included in the original work were specialized to particular tasks (e.g. summarization), and would be inapplicable for some of the tasks included in \emph{RealHall}. To handle these cases, we wrote lightly-adapted versions of the original prompts that replaced task-specific references with appropriate ones for the task being considered (e.g. QA).

When computing G-Eval, we used \texttt{gpt-3.5-turbo}, where the original paper used the now-deprecated \texttt{text-davinci-003}.  The former is generally considered a stronger model than the latter, so we expect this to work in G-Eval's favor.

When computing GPTScore, we use probabilities from \texttt{text-curie-001}. The original paper computed probabilities with many models and did not make an overall recommendation, though they noted that larger models weren't necessarily better. Our internal tests show that, across all probability-based methods we've tried, \texttt{text-curie-001} works as well or better than most OpenAI models.

\subsection{SummEval case study}\label{summeval}

We evaluated a subset of metrics on SummEval, and present results in Table \ref{summeval}. All results here were reproduced independently.

As in earlier work, we present correlation coefficients between metrics and human-annotated scores. We only consider the human-annotated Consistency score, as this is the score that captures closed-domain hallucination.

\begin{table}[H]
    \begin{tabularx}{\textwidth}{bss}
         \textbf{Metric} & $\rho$ & $\tau$  \\
        \hline
          text-curie-001 Perplexity & \textbf{0.458} & \textbf{0.364}  \\
         \hline
          GPTScore & \textbf{0.460} & \textbf{0.366} \\
         \hline
           ChainPoll-Adherence & 0.427 & 0.383 \\
         \hline
         UniEval \citep{zhong2022unified} & 0.441 & 0.354 \\
         \hline
         G-Eval & 0.309 & 0.252  \\
         \hline

    \end{tabularx}
    \caption{Spearman $\rho$ and Kendall $\tau$ correlations between metrics and human-annotated Consistency on SummEval.}
    \label{tab:summeval}
\end{table}

Our results here reproduce the strong performance for GPTScore reported in the original paper \citep{fu2023gptscore}.

At first glance, this is puzzling, since we GPTScore performed poorly on \emph{RealHall}. What accounts for the discrepancy?

To answer the question, we begin by noting that we can match GPTScore's performance simply by computing the \emph{perplexity} of the completion given the prompt, without GPTScore's added prefix.

This suggests that the strong performance of GPTScore on SummEval results from perplexity, rather than the prefix.

\begin{figure}[H]
    \centering
    \includegraphics[width=\textwidth]{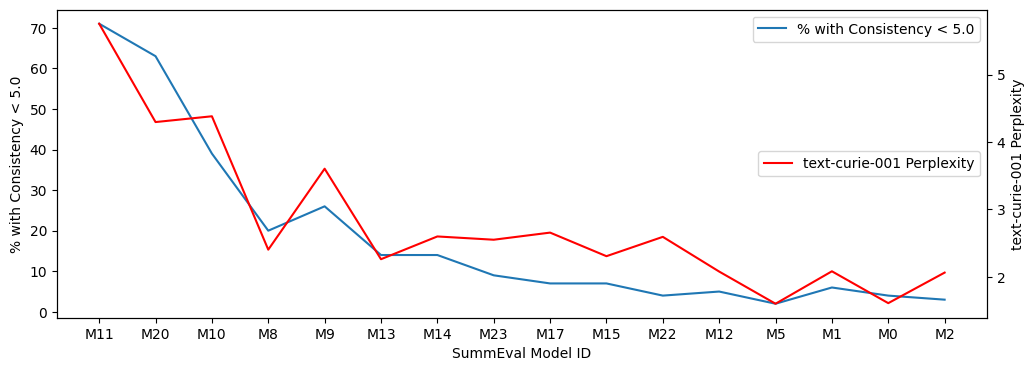}
    \caption{
    Blue: Fraction of responses from each SummEval model receiving Consistency scores less than the maximum of 5.0.
    Red: Average \texttt{text-curie-001} perplexity of each SummEval model.
    }
    \label{fig:summevalmodels}
\end{figure}

Why would perplexity perform so well on SummEval, while failing on \emph{RealHall} (as evidenced by our GPTScore results on \emph{RealHall})?  Figure \ref{fig:summevalmodels} outlines our diagnosis.

SummEval contains responses from 24 models. All of these are weaker that today's SOTA LLMs, and some are \emph{much} weaker, generating nearly incoherent text.

A large fraction of the responses that receive a less-than-perfect Consistency score come from a small subset of these under-performing models, such as \texttt{M11}.

Because modern LLMs are much better at generating coherent text, they assign high perplexity (low likelihood) to the incoherent text generated by these models. ``Detecting hallucination'' in this case only requires being able to distinguish very low-quality text -- text that a modern LLM \textbf{would be very unlikely to generate}.

To further illustrate the point, we include three example summaries from \texttt{M11}, the model at the left end of Figure \ref{fig:summevalmodels}.

It should go without saying that this type of material bears no resemblance to the much subtler cases of hallucination we need to today with today's LLMs:

\begin{quote}
    \texttt{video game `` space invaders '' was developed in japan back in 1970 . the classic video game is the latest in the u.s.-based wwe . the is the of the new japan pro wrestling organization . the `` classic game '' has been in japan 's upper house for a second stint in politics in 2013 . the former is the founder of new japan 's new japan . }

    \texttt{donald sterling , nba team last year . sterling 's wife sued for \$ 2.6 million in gifts . sterling says he is the former female companion who has lost the . sterling has ordered v. stiviano to pay back \$ 2.6 m in gifts after his wife sued . sterling also includes a \$ 391 easter bunny costume , \$ 299 and a \$ 299 .}

    \texttt{foxes host swansea on saturday just three points from the premier league . nigel pearson has urged leicester to keep their cool and ignore their relegation rivals . jamie vardy scored an injury-time winner against west bromwich albion on saturday . the foxes host the foxes at west brom in sunday .}    
\end{quote}

\subsection{Detailed results} \label{detailedresults}

\vspace{2pt}
\begin{table}[H]
    \begin{tabularx}{\textwidth}{bb}
        \hline
         \textbf{Metric} & \textbf{AUROC}  \\
         \hline
\emph{ChainPoll-Correctness} & \mybarbold{0.697}{0.394}{6}{154}{243} \\ 
\hline

\emph{ChainPoll-Correctness} (w/o detailed CoT) & \mybar{0.629}{0.258}{177}{196}{75} \\ 
\hline

SelfCheck-Bertscore & \mybar{0.555}{0.110}{255}{150}{0} \\ 
\hline

SelfCheck-NGram & \mybar{0.516}{0.031}{255}{91}{0} \\ 
\hline

G-Eval & \mybar{0.533}{0.067}{255}{118}{0} \\ 
\hline

Max pseudo-entropy & \mybar{0.520}{0.040}{255}{98}{0} \\ 
\hline

GPTScore & \mybar{0.487}{-0.026}{255}{69}{0} \\ 
\hline

Random Guessing & \mybar{0.500}{0.000}{255}{69}{0} \\ 
\hline
    \end{tabularx}
    \caption{AUROC scores for open-domain hallucination detection on Open Assistant Prompts.}
    \label{tab:oasst}
\end{table}

\begin{table}[H]
    \begin{tabularx}{\textwidth}{bb}
        \hline
         \textbf{Metric} & \textbf{AUROC}  \\
         \hline

           \emph{ChainPoll-Correctness} & \mybarbold{0.847}{0.694}{6}{154}{243} \\ 
\hline

\emph{ChainPoll-Correctness} (w/o detailed CoT) & \mybar{0.818}{0.635}{47}{164}{202} \\ 
\hline

SelfCheck-Bertscore & \mybar{0.784}{0.569}{95}{176}{155} \\ 
\hline

SelfCheck-NGram & \mybar{0.757}{0.513}{134}{185}{117} \\ 
\hline

G-Eval & \mybar{0.615}{0.230}{255}{165}{0} \\ 
\hline

Max pseudo-entropy & \mybar{0.611}{0.221}{255}{161}{0} \\ 
\hline

GPTScore & \mybar{0.492}{-0.017}{255}{69}{0} \\ 
\hline

Random Guessing & \mybar{0.500}{0.000}{255}{69}{0} \\ 
\hline

               \end{tabularx}
    \caption{AUROC scores for open-domain hallucination detection on TriviaQA.}
    \label{tab:tqa}
\end{table}

\begin{table}[H]
    \begin{tabularx}{\textwidth}{bb}
        \hline
         \textbf{Metric} & \textbf{AUROC}  \\
         \hline
\emph{ChainPoll-Adherence} & \mybarbold{0.785}{0.569}{6}{154}{243} \\ 
\hline

\emph{ChainPoll-Adherence} (w/o detailed CoT) & \mybar{0.707}{0.413}{142}{187}{109} \\ 
\hline

SelfCheck-Bertscore & \mybar{0.686}{0.372}{177}{196}{75} \\ 
\hline

SelfCheck-NGram & \mybar{0.581}{0.161}{255}{151}{0} \\ 
\hline

TRUE & \mybar{0.727}{0.454}{105}{178}{145} \\ 
\hline

G-Eval & \mybar{0.650}{0.300}{240}{211}{14} \\ 
\hline

Max pseudo-entropy & \mybar{0.552}{0.104}{255}{121}{0} \\ 
\hline

GPTScore & \mybar{0.622}{0.244}{255}{193}{0} \\ 
\hline

Random Guessing & \mybar{0.500}{0.000}{255}{69}{0} \\ 
\hline
    \end{tabularx}
    \caption{AUROC scores for closed-domain hallucination detection on CovidQA.}
    \label{tab:covidqa}
\end{table}

\begin{table}[H]
    \begin{tabularx}{\textwidth}{bb}
        \hline
         \textbf{Metric} & \textbf{AUROC}  \\
         \hline
\emph{ChainPoll-Adherence} & \mybarbold{0.794}{0.589}{6}{154}{243} \\ 
\hline

\emph{ChainPoll-Adherence} (w/o detailed CoT) & \mybar{0.537}{0.074}{255}{105}{0} \\ 
\hline

SelfCheck-Bertscore & \mybar{0.665}{0.329}{224}{207}{29} \\ 
\hline

SelfCheck-NGram & \mybar{0.722}{0.445}{127}{183}{124} \\ 
\hline

TRUE & \mybar{0.459}{-0.082}{255}{69}{0} \\ 
\hline

G-Eval & \mybar{0.517}{0.035}{255}{86}{0} \\ 
\hline

Max pseudo-entropy & \mybar{0.519}{0.037}{255}{87}{0} \\ 
\hline

GPTScore & \mybar{0.494}{-0.013}{255}{69}{0} \\ 
\hline

Random Guessing & \mybar{0.500}{0.000}{255}{69}{0} \\ 
\hline
    \end{tabularx}
    \caption{AUROC scores for closed-domain hallucination detection on DROP. The dramatic gap between the ChainPoll and ChainPoll (without detailed CoT) reflects chain-of-thought prompting on discrete reasoning tasks.}
    \label{tab:drop}
\end{table}

\end{document}